\pdfoutput=1

\documentclass[10pt,twocolumn,letterpaper]{article}

\usepackage{cvpr}
\usepackage{times}
\usepackage{epsfig}
\usepackage{graphicx}
\usepackage{amsmath}
\usepackage{amssymb}
\usepackage[breaklinks=true,bookmarks=false]{hyperref}

\cvprfinalcopy % *** Uncomment this line for the final submission

\def\cvprPaperID{****} % *** Enter the CVPR Paper ID here

\begin{document}

%%%%%%%%% TITLE
\title{Cultivating DNN Diversity for Large Scale Video Labelling}

%\author[]{Mikel Bober-Irizar}
%\author[]{Sameed Husain}
%\author[]{Eng-Jon Ong}
%\author[]{Miroslaw Bober}
%\affil[]{CVSSP, University of Surrey, UK}
%\affil[T]{Visual Atoms Ltd.}

\author{Mikel Bober-Irizar \thanks{The author is with CVSSP, University of Surrey, UK}\\
%Visual Atoms\\
%\thanks{The author is with Visual Atoms Ltd, }\\
%Institution1 address\\
{\tt\small mikel@mxbi.net}
% For a paper whose authors are all at the same institution,
% omit the following lines up until the closing ``}''.
% Additional authors and addresses can be added with ``\and'',
% just like the second author.
% To save space, use either the email address or home page, not both
%\and
\and
Sameed Husain $^{*}$\\
{\tt\small sameed.husain@surrey.ac.uk}\\
\and
Eng-Jon Ong $^{*}$\\
{\tt\small e.ong@surrey.ac.uk}\\
\and
Miroslaw Bober $^{*}$
\thanks{The author is also with Visual Atoms Ltd, Guildford, UK}\\
{\tt\small m.bober@surrey.ac.uk}\\
}

\date{CVSSP, University of Surrey, UK}

\maketitle
%\thispagestyle{empty}
%%%%%%%%% ABSTRACT
\begin{abstract} 
We investigate factors controlling DNN diversity in the context of the ``Google Cloud and YouTube-8M Video Understanding Challenge''. While it is well-known that ensemble methods improve prediction performance, and that combining accurate but diverse predictors helps, there is little knowledge on how to best promote $\&$ measure DNN diversity. We  show that diversity can be cultivated by some unexpected means, such as model over-fitting or dropout variations. We also present details of our solution to the video understanding problem, which ranked $\#$7 in the Kaggle competition (competing as the Yeti team).
\end{abstract}

\pdfoutput=1

\section{Introduction}
Accurate clip-level video classification, utilising a rich vocabulary of sophisticated terms, remains a challenging problem.
One of the contributing factors is the complexity and ambiguity of the interrelations between linguistic terms and the actual audio-visual content of the video. For example, while a "travel" video can depict any location with any accompanying sound, it is the {\em intent of the producer} or even the {\em perception of the viewer} that makes it a "travel" video, as opposed to a "news" or "real estate" clip. Hence true {\em understanding} of the video's meaning is called for, and not mere {\em recognition} of the constituent locations, objects or sounds. 

%Another difficulty is that large volumes of consistently labelled data are difficult to source. 
%The excuse of the "missing training data" has been recently removed by Google Cloud organising a Kaggle competition on "Google Cloud & YouTube-8M Video Understanding Challenge".

The recent Kaggle competition entitled "Google Cloud \& YouTube-8M Video Understanding Challenge" provides a unique platform to benchmark existing methods and to develop new approaches to video analysis and classification. It is based around the {\bf YouTube-8M (v.2)} dataset, which contains approximately 7 million individual videos, corresponding to almost half a million hours (50 years!), annotated with a rich vocabulary of 4716 labels \cite{45619}. The challenge for participants was to develop classification algorithms which accurately assign video-level labels. 

Given the complexity of the video understanding task, where humans are known to use diverse clues, we hypothesise that a successful solution must efficiently combine different expert models. We pose two important questions: (i) How do we construct such diverse models and how to combine them?, and (ii) do we need to individually train and combine discrete models or can we simply train a very large/flexible DNN to obtain a fully trained end-to-end solution? The first question clearly links to ensemble-based classifiers, where significant body of prior work demonstrates that diversity is important. However, do we know all the different ways to promote diversity in DNN architectures? On the second question, our analysis shows that training a single network results in sub-optimal solutions as compared to an ensamble.

In the following section we briefly review the state-of-the-art in video labelling and ensemble-based classifiers. We then introduce the Kaggle competition, including data-sets, performance measures and the additional features engineered and evaluated by the Yeti team. Next, in Section \ref{sec:dnns}, we describe the different forms of DNNs that were employed and quote the baseline performance of individual DNNs trained on different features. Section \ref{sec:exp} demonstrates that further gains in performance can be achieved by promoting diversification of DNNs during training by adjusting dropout rates, different architectures and - surprisingly - using over-fitted DNNs. We then provide analysis on the link between diversity of the DNNs in the final {\em Yeti} ensemble, performance gains in Section \ref{sec:diverse_analysis}, and conclude in Section \ref{sec:conclusions}.

\section{Related Work}
We first overview some existing approaches to video classification before discussing ensemble-based classifiers. Ng et al. \cite{7299101} introduced two methods which aggregate frame-level features into video-level predictions: Long short-term memory (LSTM) and Feature pooling. Fernando et al. \cite{7458903} proposed a novel rank-based pooling method that captures the latent structure of the video sequence data. 
Karpathy et al. \cite{6909619} investigated several methods for fusing information across temporal domain and introduced Multiresolution CNNs for efficient video-classification. 
Wu et al. \cite{Wu} developed a multi-stream architecture to model short-term motion, spatial and audio information respectively. LSTM are then used to capture long-term temporal dynamics. 

DNNs are known to provide significant improvement in performance over traditional classifiers across a wide range of datasets. However, it was also found that further significant gains can be achieved by constructing ensembles of DNNs. One example is the ImageNet Large Scale Visual Recognition Challenge (ILSVRC) \cite{ILSVRC15}. Here, improvements up to 5\% were achieved over individual DNN performance (e.g. GoogLeNet\cite{googlenet}) by using ensembles of existing networks. Furthermore, all the top entries in this challenge employed ensembles of some form.

One of the key reasons for such a large improvement was found to be due to the diversity present across different base classifiers (i.e. different classifiers specialise to different data or label subsets)\cite{Hansen,Krogh95}. An increase in diversity of classifiers of equal performance will usually increase the ensemble performance. There are numerous methods for achieving this: random initialisation of the same models, or data modification using Bagging \cite{Breiman} or Boosting \cite{Boosting} processes. Recently, work was carried out on end-to-end training of an ensemble based on diversity-aware loss functions. Chen et al. \cite{NCL} proposed to use Negative Correlation Learning for promoting diversity in an ensemble of DNNs, where a penalty term based on the covariance of classifier outputs is added to an existing loss function. An alternative was proposed by Lee et al \cite{Mheads} based on the approach of Multiple Choice Learning (MCL) \cite{MCL}. Here, DNNs are trained based on a loss function that uses the final prediction chosen from an individual DNN with the lowest independent loss value.

\section{Youtube-8M Kaggle competition}
\label{sec:data}
The complete Youtube-8M dataset consists of approximately 7 million Youtube videos, each approximately 2-5 minutes in length, with at least 1000 views each. There are 4716 possible classes for each video, given in a multi-label form. For the Kaggle challenge, we were provided with 6.3 million labelled videos (i.e. each video was associated with a 4716 binary vector for labels). For test purposes, approximately 700K unlabelled videos were provided. The resulting class test predictions from our trained models were uploaded to the Kaggle website for evaluation. 

The evaluation measure used is called `GAP20'. This is essentially the mean average precision of the top-20 ranked predictions across all examples. To calculate its value, the top-20 predictions (and their corresponding ground-truth labels) are extracted for each test video. The set of top-20 predictions for all videos is concatenated into a long list of predictions. A similar process is performed for the corresponding ground-truth labels. Both lists are then sorted according to the confidence prediction values and mean average precision is calculated on the resulting list.

Below we present different features used for classification; the first two (FF, MF) were provided by Google, the remaining ones were computed by our team. For some features, we also quote the performance as a rough guide of the usefulness of the individual feature: this was computed by 4k-4k-4k DNN with dropout $0.4$.

\begin{figure}[!t]
 	\centering
 	\includegraphics[width=\columnwidth]{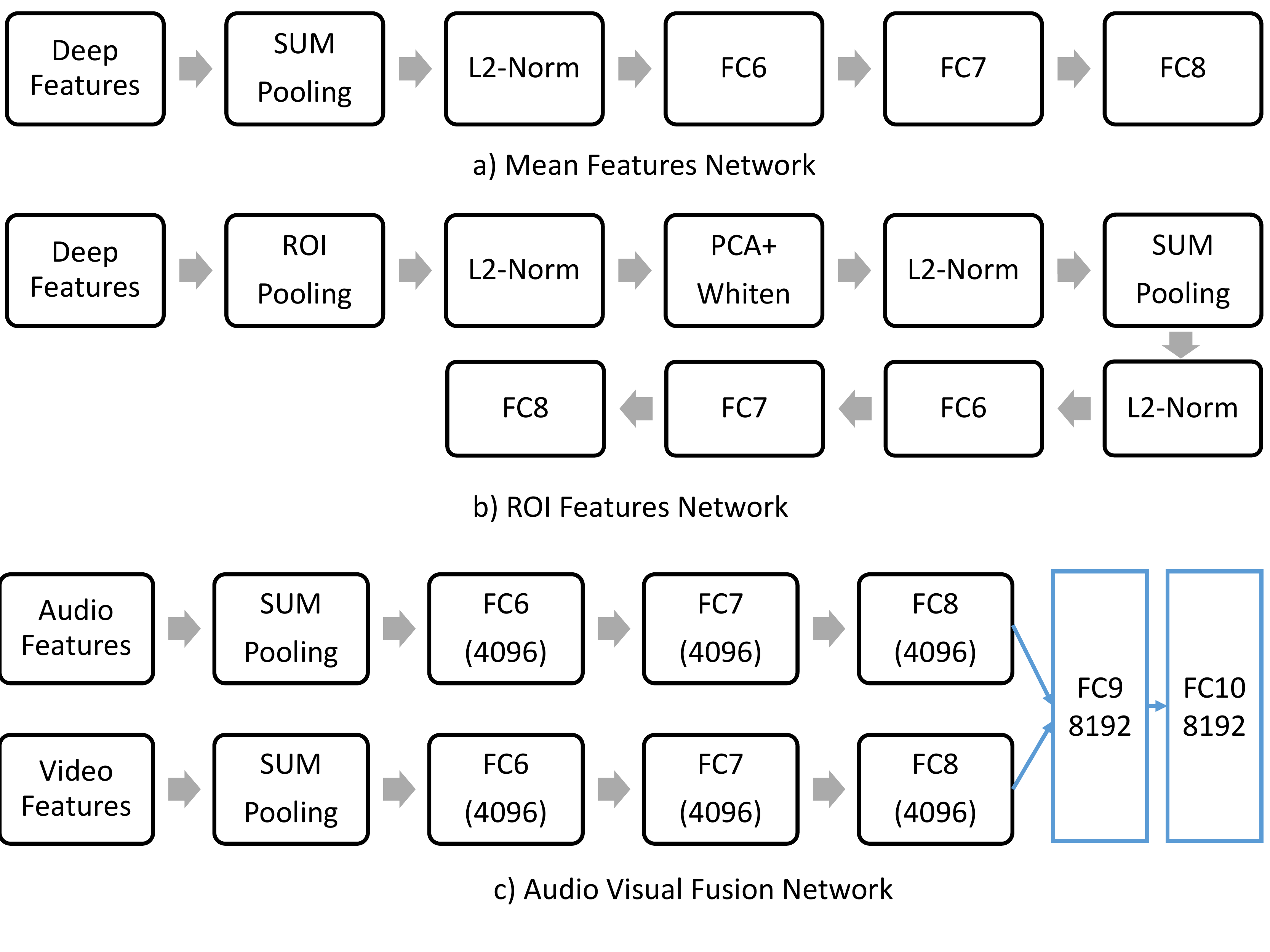}
 	\caption[\bf{CNN architectures}]{(a) Mean features CNN (b) ROI features CNN (c) Audio Visual fusion CNN.}
 	\label{GDES}
 \end{figure}

\begin{itemize}

\item{\bf Frame-level Deep Features (FL)}\\
In the Kaggle challenge, the raw frames (image data) of the videos were not provided. Instead, each video in the dataset was decoded at 1 frame-per-second up to first 300 seconds and then passed through an Inception-v3 network \cite{44903}. The ReLu activation values of the last hidden layer formed a frame-level representation (2048 dimensions), which was subsequently reduced to 1024 dimensions using a PCA transformation with whitening. Similar processing was performed on the audio stream, resulting in an additional 128-dimensional audio feature vector. Video and audio features are concatenated to yield a frame feature vector of 1152 dimensions. The set of frame-level deep features extracted for a video $I$ is denoted as $\mathcal X^{I}=\{x_{t}\in \mathbb{R}^{d}, t=1...T\}$.

\end{itemize}
The extracted features are then aggreagted using state-of-the-art aggregation methods: Mean aggregation, Mean + Standard Deviation aggregation, ROI-Pooling \cite{Radenovic2016CNNIR}, VLAD, Fisher Vectors \cite{Jegou12PAMI}, RVD \cite{RVD} \cite{BRVD} \cite{RVD_PAMI} and BoW.
\begin{itemize}

\item{\bf Video-Level Mean Features (MF)}\\
 Google also provided the mean feature $\mu^I$ for each video, which was obtained by averaging frame-level features across the time dimension.
 Our reference performance for MF feature is $81.94\%$, but it can peak at $82.55\%$ with a 12k-12k-12k network and dropout of $0.4$.

\item{\bf Video-Level Mean Features + Standard Deviation (MF+STD)}\\
We extract the standard deviation feature $\sigma^I$ from each video. The signature $\sigma^I$ is L2-normalised and concatenated with mean feature $\mu^I$ to form a 2304-Dim representation $\phi^I=[\mu^I;\sigma^I]$.

\item{\bf Region of Interest pooling (ROI)}\\
The ROI-pooling based descriptor, proposed by Tolias et al \cite{Radenovic2016CNNIR}, is a global image representation that achieves state-of-the-art performance in image retrieval and classification. We compute a new video-level representation using the ROI-pooling approach. More precisely, the frame-level features are max-pooled across 10 temporal-scale overlapping regions, obtained from a rigid-grid covering the frame-level features, producing a single signature per region. These region-level signature are independently L2-normalised, PCA transformed and subsequently whitened. The transformed vectors are then sum-aggregated and finally L2-normalised. The dimensionality of final video-level representation is 1152. The ROI-based training architecture is presented in Fig. \ref{GDES}(b); it achieves $82.34\%$ with the 12k-12k-12k net. 

\item{\bf Fisher Vectors, RVD and VLAD pooling (FV, RVD, VLAD)}\\
We encode the frame level features using the classical Fisher Vectors, RVD and VLAD approaches. Fisher Vector encoding aggregates local features based on the Fisher Kernel framework while VLAD is simplified version of Fisher vectors. The detailed experimental results show that the mean pooling achieves significantly better classification accuracy than FV, RVD and VLAD approaches (81.94\%  vs 81.3\%, 80.8\% and 80.4\%) [4k-4k-4k network].

\item{\bf BoW pooling (BoW)}\\
We compute the BoW representation of the frame-level video features, using 2k and 10k BOW representations. We compute BoW features by first applying K-Means clustering across the frame-level deep features with either 2k or 10k clusters, and then calculating the number of frames in each cluster for each video. Finally, we L1-normalize this BoW vector to remove the effect of video length on the features. The base BoW performance is $78.1\%$ with the 4k-4k-4k net. 

\end{itemize}

\section{DNN-based Multi-Label Classifiers}
\label{sec:dnns}
This section describes the base neural network architectures that we used for multi-label predictions on this dataset.
%Additionally, we also detail the features that were used as inputs to the NNs.

\subsection{Fully Connected NN Architecture}
For our work, we use a 3-hidden layer fully connected neural network, with layers FC6, FC7 and FC8. The size of the input layer is dependent on the feature vectors chosen. These will be described in more detail in the sections below.

The activation function for all hidden units is ReLU. Additionally, we also employ dropout on each hidden layer.
The number of hidden units and dropout value will be detailed in Section \ref{sec:exp}.

The output layer is again a fully connected layer, with a total of 4716 output units, one for each class. In order to provide class-prediction probability values, each output unit uses a sigmoid activation function.

Since this challenge is a multi-label problem, we used the binary cross entropy loss function for training the DNNs. We have chosen the Adam optimization algorithm \cite{adam} for training, with a learning rate of $1e-4$. Using the learning rate of $1e-3$ as in the original paper led to NNs getting stuck in local minima very early in training.
All the DNNs were trained to convergence, and the number of epochs required to achieve this ranged from 15 to 150 depending on the hyper-parameter settings chosen, as detailed in Section \ref{sec:exp}. 

\subsection{Audio Visual Fusion}
This method comprises of two stages: Firstly the audio and visual features networks are trained separately to minimise the classification loss and then these two NNs are integrated in a fusion network consisting of two fully-connected layers.

1) Training audio and video networks: We first train the audio and video networks individually. This is conducted by connecting their features to three fully connected layers similar to FC6, FC7 and FC8, respectively. The size of all FC layers is 4096. Each FC layer is followed by a ReLu and a dropout layer. The output of the FC8 layer is passed through another fully connected layer FC9 which computes the predictions and finally updates the network parameters to minimise the cross entropy loss over the training data.

2) Training fusion networks: After training the audio and video networks, we discard their FC9 layers and connect their FC8 layers to the fusion network shown in Fig. \ref{GDES}(c). In this way, 4096-Dim audio and 4096-Dim video features are concatenated to form a 8192-Dim representation as an input to the fusion network. This fusion network contains two fully connected layers of size 8192, followed by a fully connected prediction layer and cross-entropy optimisation.

The model based on Audio-Visual Fusion achieved $82.28\%$, but added significant diversity to our ensemble. 

\subsection{Ensemble of DNNs}
The test predictions from multiple DNNs that were trained separately with different architectures, input features and hyper-parameters can be combined together by averaging them. We have found such ensembles often provide significant improvements over the performance of individual DNNs. The details of the diversification process and ensemble construction is presented in Section \ref{sec:exp}.

\section{Diversification of DNNs}
\label{sec:exp}
It was found that the performance of individual models (architectures and input features) could be significantly improved when they were combined together into a DNN ensemble. However, in order to achieve these gains, it was necessary to build diverse DNNs. In this section, we describe a number of approaches that we have attempted that were mostly successful in achieving the aim of diversification of DNNs. These range from using different dropouts, hidden unit counts, use of overfitted models and segmented frame-level features. 
%In order to analyse the diversity of the ensemble, we first need to quantify it.

\subsection{Sizes of Hidden Layers}
For our experiments, we have considered the use of the following number of units for each hidden layers: 4096, 8192, 10240, 12288 and 16384. All the hidden layers within a model were set to have the same number of hidden units, as we did not see substantial gains by varying the hidden layer size within a model.

\subsection{Dropout Sizes}
In the process of training different DNNs, a number of different dropout values were used: 0.0 (no dropout), 0.25, 0.3, 0.4 and 0.5. As expected, we have found that the higher the dropout, the larger the number of epochs required for convergence to be achieved.
% TODO - why use this dropout?

\subsection{Use of Overfitted DNNs}
We have also used models that are overfitting. We have found that individual DNNs will have a validation GAP20 score that peaks after a certain number of training epochs (usually around 40-50 for large networks of ${>}$8K units). If training continues, we find that the validation GAP20 score will steadily decrease. This implies that the model is overfitting to the training data. Existing practise often discards these models and use the model with the best validation score. 
Counter-intuitively, using large network models that have overfitted was found to give a {\em larger} performance improvement to the ensemble of classifiers. This is despite individual validation GAP scores that are less than its peak GAP score many epochs prior.

\subsection{Using Different Training Subsets}
Finally, we have explored how using different training subsets for building similar architectures can influence the final performance of the ensemble. To this end, we first trained a DNN ensemble using DNNs with the above different hidden units, dropouts and feature vectors (ROI, video mean, video mean + std. dev.) using the training dataset and validation set (except last 100K validation data) provided by Google.

We then produced another set of training data and 100K validation data that was split differently to the above. Next, DNNs with 4K-4K-4K, 4K-8K-8K, 8K-8K-8K, 10K-10K-10K, 12K-12K-12K, 14K-14K-14K and 16K-16K-16K architectures were trained using a separate training set for the video-mean features. These were used to form a separate DNN ensemble. We have found that this also provides improvement when outputs of both ensembles are linearly combined together.

\subsection{Diversification-based Loss Function}
\label{sec:divbased}
Recently, there has been work on performing end-to-end learning of multiple DNN output layers that promote diversity \cite{Opitz2017} for the task of multi-class classification. Here, a multi-output layer DNN was proposed. The final output label is the class with the maximum votes from all the output layers. In order to learn this DNN, a ``diversity-aware'' loss function was proposed. This was a linear combination of the MSE error with the sum of cross-entropy of outputs for the different layers. The aim was not only to have each output layer minimise classification error, but to also provide classification outputs that are different from other output layers.

We have attempted to use a similar approach to sequentially train diversity-aware DNNs. In order to achieve this, we first train a single 3-layer fully connected DNN as described above. Our ensemble is initialised using this single DNN. The outputs for the training data of this ensemble is then recorded for subsequent use.

In order to further add new DNNs into our ensemble, we wish to learn DNNs that minimise labelling errors {\em and} produce outputs that are different to that of the ensemble.
Learning the next DNNs was performed by proposing a loss function that accounts for multi-label classification accuracy and is also diversity-aware. For the multi-label classification, we have used the binary cross entropy method. For the diversity awareness, we use the negative of cross entropy between the current DNN and the ensemble output. The final loss function is a linear combination between the above two losses, with a combination parameter of $\lambda = 0.3$ for diversity and $0.7$ for multi-label accuracy. The new DNN is then added into the ensemble set and this step repeated a number of times until a pre-defined ensemble size is reached (here we chose 4).

\section{Experimental Results}
In this section, we shall provide results from the individual models. We also show how a significantly improved GAP20 score can be achieved by combining the individual classifiers into an ensemble. We further achieve improvement by performing linear combination on ensembles trained using different training sets.

\subsection{Performance of Individual Features}
In this section, we detail the baseline performances of DNNs that were trained on the different features described above.
% TODO - table, and brief statement of the results from the table.
\begin{table}[h!]
\centering
\begin{tabular}{|c c c c c|} 
 \hline
 Features & Architecture & DropOut & GAP20 & Peak \\ 
 \hline \hline
 Mean & 4k-4k-4k & 0.25 & 81.94 & 81.94 \\ 
 \hline
 Mean & 4k-4k-4k & 0.3 & 82.01 & 82.01 \\
 \hline
 Mean &	4k-4k-4k &	0.4	& 82.08	& 82.08\\
 \hline
 Mean &	8k-8k-8k &	0.4 &	82.48 &	82.48\\
 \hline
 Mean &	12k-12k-12k &	0.4	 & 82.31 &	82.55\\
 \hline
 ROI	 & 8k-8k-8k	& 0.4	& 82.25 &	82.25\\
 \hline
 ROI	& 12k-12k-12k &	0.4	 & 82.12 &	82.34\\
 \hline
 Fusion &	8k-8k-8k &	0.4 &	82.28 &	82.28\\
 \hline
 Mean+sd &	8k-8k-8k &	0.3 &	82.1 &	82.1\\
 \hline
 Mean+sd &	10k-10k-10k &	0.4 &	82.54 &	82.54\\
 \hline
 Mean+sd &	12k-12k-12k &	0.4 &	82.41 &	82.6\\
 \hline
\end{tabular}
\bigskip
\caption{Table showing the GAP performance of different architectures for the first ensemble, features and dropout settings. Shown are two GAP scores, one at the last epoch (GAP20) and another at the epoch where the peak GAP score was achieved. }
\label{table:indiv_perf}
\end{table}

It can be observed from the Table \ref{table:indiv_perf} that GAP20 score increases as we increase the dropouts percentage, keeping the rest of the network hyperparameters the same. Also the deeper 8k-8k-8k architecture performs significantly better than 4k-4k-4k using dropout 0.4. Furthermore, adding second order statistics (standard deviation features) to mean features increases the GAP20 from 82.0\% to 82.1\%. The ROI and Fusion CNNs performs marginally less that Mean CNNs. However, all the architectures presented add value to the overall performance.  

\begin{table}[h!]
\centering
\begin{tabular}{|c c c c c|} 
 \hline
 Features & Architecture & DropOut & GAP20 & Peak \\ 
 \hline \hline
 Mean & 4k-4k-4k & 0.25 & 81.87 & 81.87 \\ 
 \hline
 Mean & 4k-4k-4k & 0.3 & 81.92 & 81.92 \\
 \hline
 Mean &	4k-4k-4k &	0.4	& 82.01	& 82.01\\
 \hline
 Mean &	8k-8k-8k &	0.4 &	82.15 &	82.48\\
 \hline
 Mean &	10k-10k-10k &	0.4	 & 82.18 &	82.28\\
 \hline
 Mean &	12k-12k-12k &	0.4	 & 82.11 &	82.38\\
 \hline
 Mean &	14k-14k-14k &	0.4	 & 82.16 &	82.24\\
 \hline
 Mean &	16k-16k-16k &	0.4	 & 82.20 &	82.41\\
 \hline
 Mean+sd &	10k-10k-10k &	0.4	 & 82.39 &	82.43\\
 \hline
 Mean+sd &	12k-12k-12k &	0.4	 & 82.37 &	82.45\\
 \hline
\end{tabular}
\bigskip
\caption{Table showing the GAP performance of different architectures for the second ensemble, trained with a different train-validation split, input features and dropout settings. Shown are two GAP scores, one at the last epoch (GAP20) and another at the epoch where the peak GAP score was achieved. }
\label{table:indiv_perf2}
\end{table}

\subsection{Performance of DNN Ensemble}
\begin{figure}[!t]
 	\centering
 	\includegraphics[width=\columnwidth]{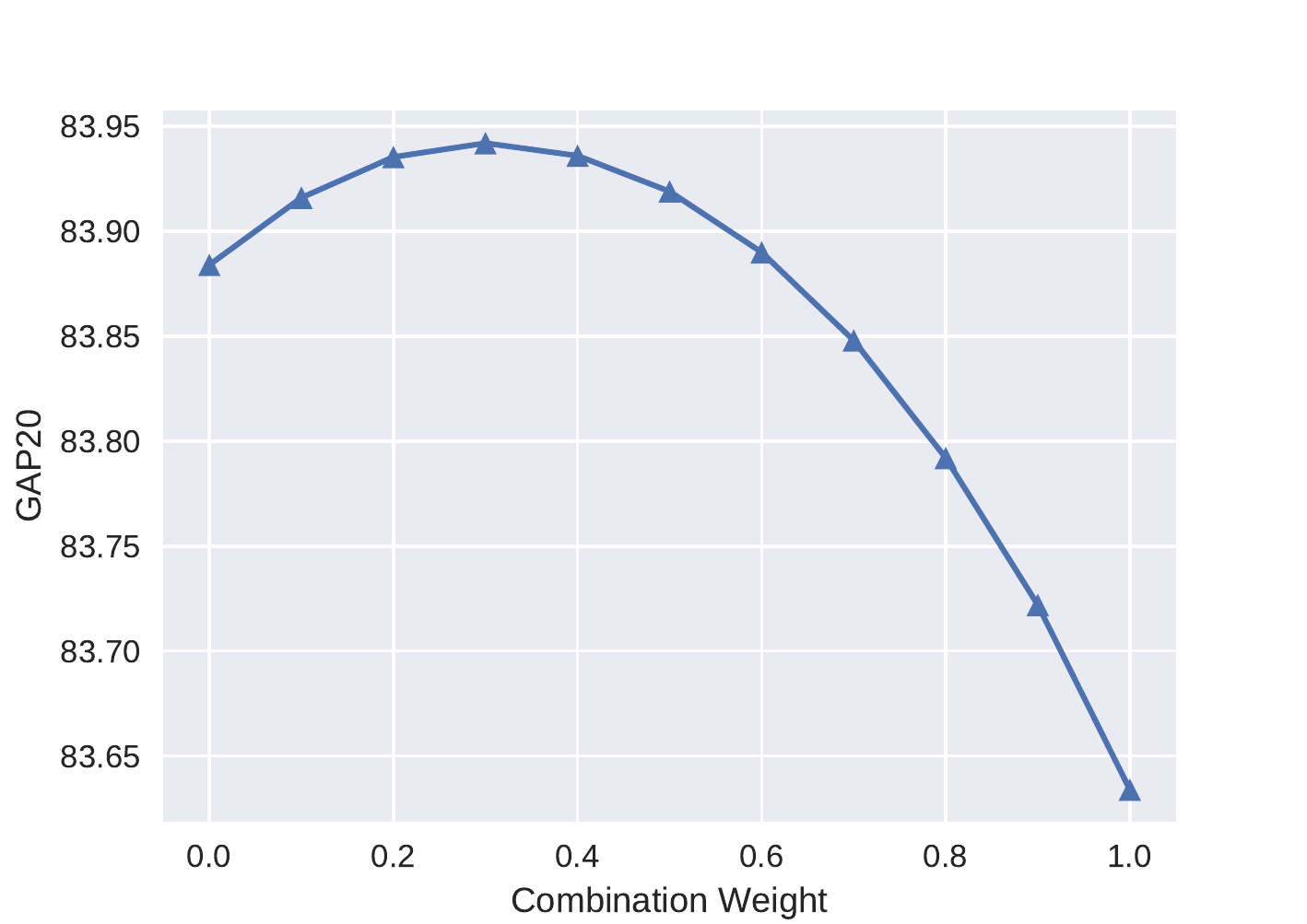}
 	\caption{ This figure shows different linear combination values for combining the two ensembles trained with different training-validation splits.}
 	\label{fig:weight_res}
 \end{figure}

We have found that the overall GAP20 performance of the ensemble $E1$ formed in Table \ref{table:indiv_perf} was 83.884\% and ensemble $E2$ from Table \ref{table:indiv_perf2} was 83.634\% on the Kaggle leaderboard. When combined together, we have found potential improvements in the linearly weighted predictions from both ensembles, with a weighting of $\alpha \in (0,1)$ for one ensemble and $1-\alpha$ for the other ensemble. The results can be seen in Fig. \ref{fig:weight_res}. The optimal GAP20 score achieved is 83.96\% on the Kaggle leaderboard using the formula $0.65\times E1+0.35\times E2$.  

An interesting discovery was that the use of overfitted DNNs can improve the generalisation performance when incorporated into an ensemble. We have found that for large DNNs, (8K and above hidden units), when models trained up to later epochs (100+) were used, the validation error of the ensemble further decreases. This is despite the {\em increase} of validation error in the individual models. We have found that the use of overfitted models resulted in an average of 0.671\% improvement in the ensemble GAP20 score, compared with 0.579\% when using peak-validation GAP models. One hypothesis is that the overfitted models are overfitting to different video and label subsets. This in turn promotes diversity across different DNNs used, which results in better generalisation of the ensemble.

The ensemble that was trained using the sequential addition with the diversity aware loss function (Section \ref{sec:divbased}) did not yield any improvement over a simple average of randomly initialised and different architecture DNNs. We found that a 4-DNN ensemble (8K-8K-8K DNN)  of learnt this way yielded a GAP score of 82.15\% and this did not improve by adding more DNNs.

\section{DNN Ensemble Diversity Analysis}
\label{sec:diverse_analysis}
It is generally agreed that greater output diversity of member classifiers in an ensemble result in improved performance. Unfortunately, the measurement of diversity is not straightforward, and at present, a generally accepted formulation does not exist \cite{Kuncheva}. Here, at least 10 different measures of diversity were found. 

For our purposes, first suppose there are $M$ classifiers in our ensemble. There are 2 diversity measures that are relevant to our analysis. The first is based on the Pearsons correlation coefficient and the second based on Generalised Diversity Measure.

\begin{figure}[!h]
 	\centering
 	\includegraphics[width=0.8\columnwidth]{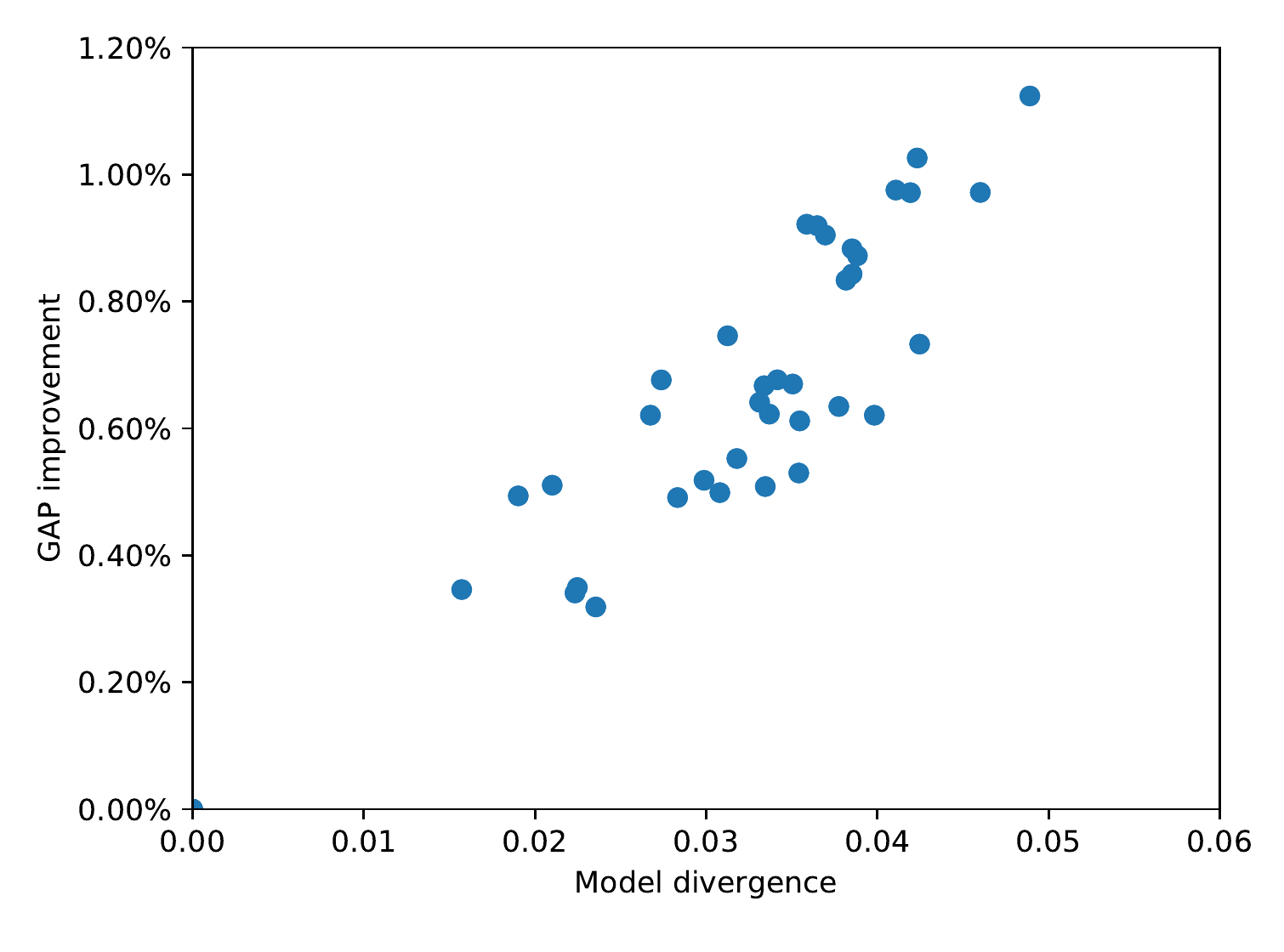} \\
 	(a)\\
 	\includegraphics[width=0.8\columnwidth]{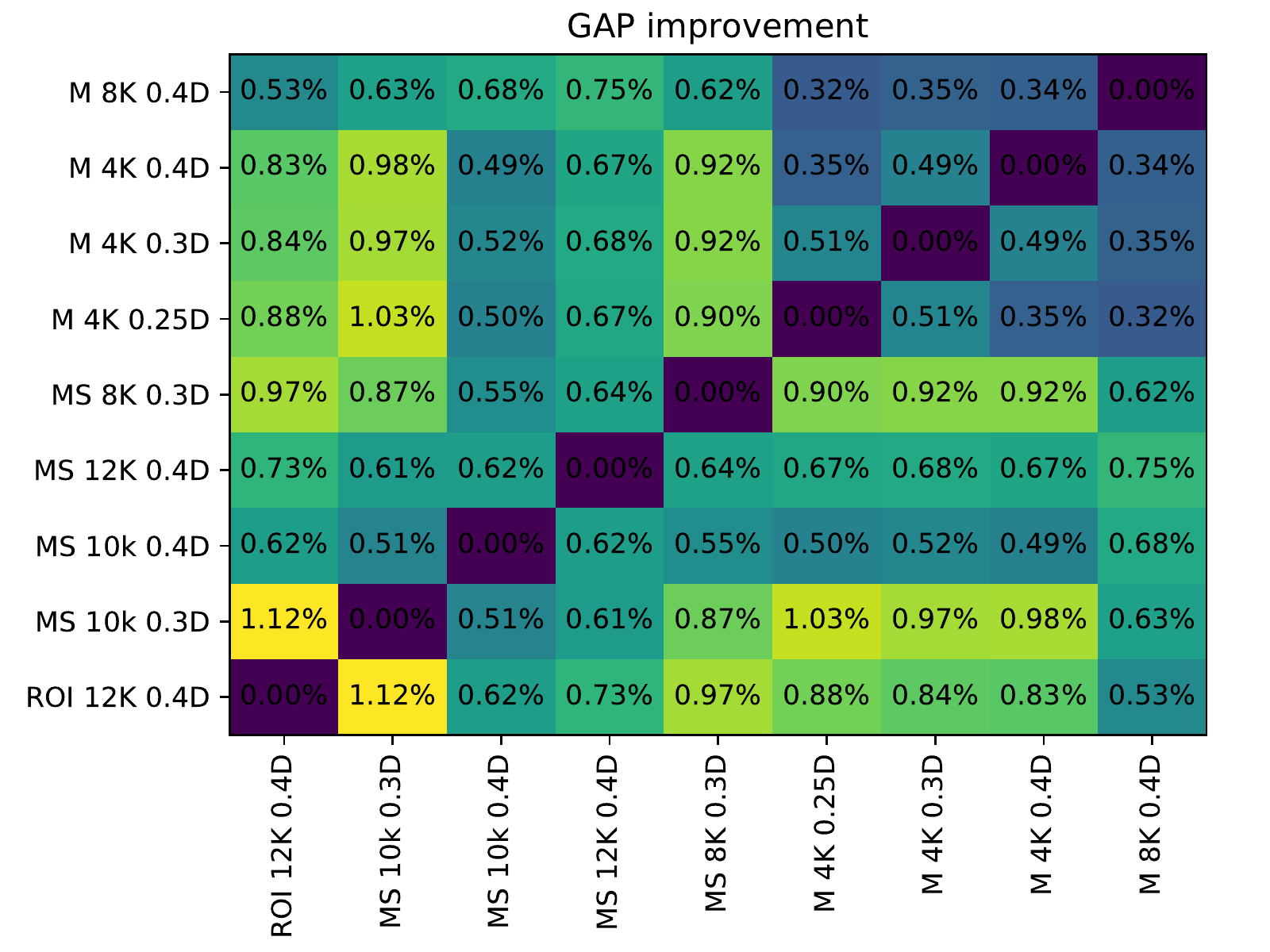} \\
 	(b)\\
 	\includegraphics[width=0.8\columnwidth]{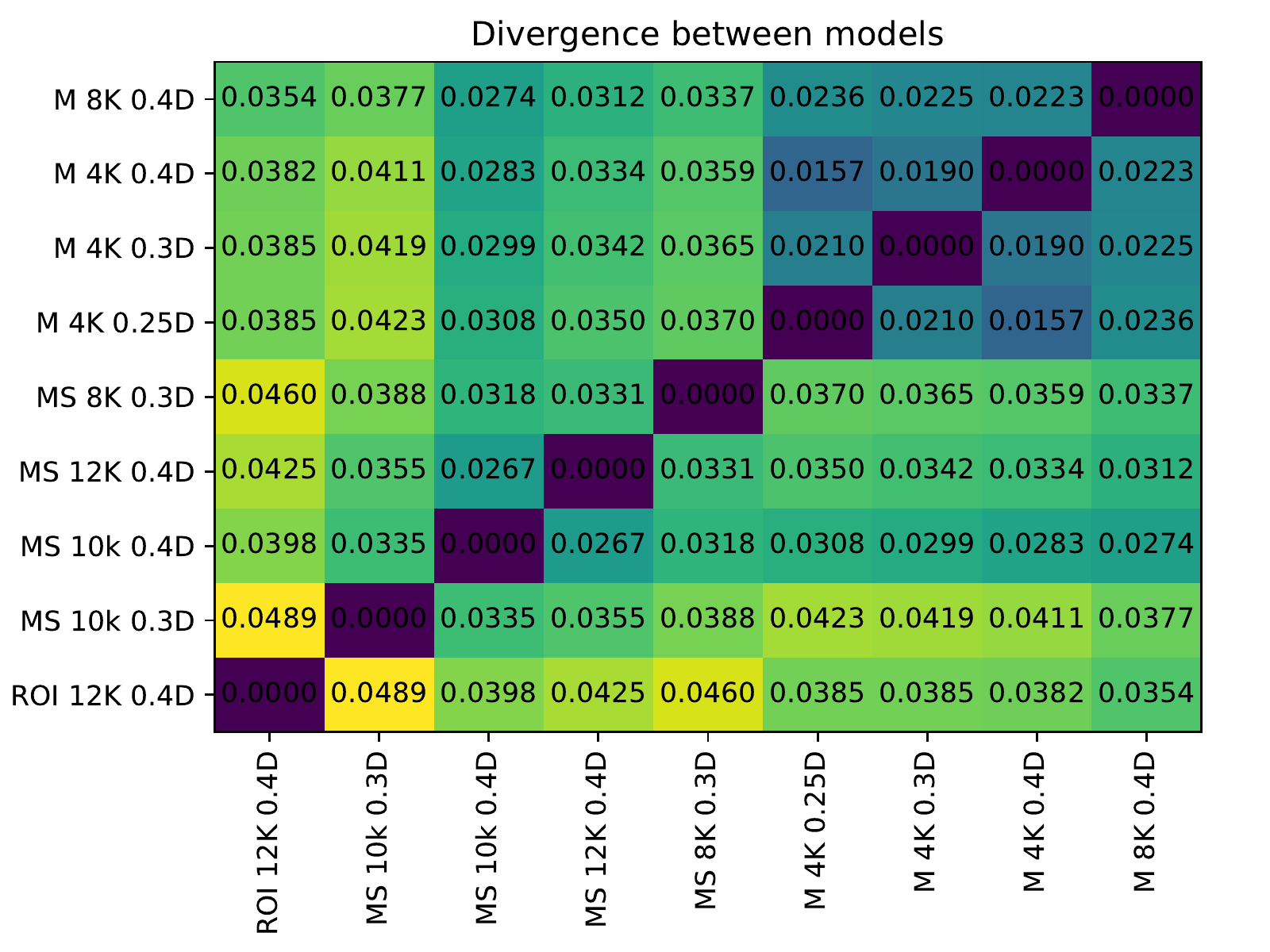} \\
 	(c)
 	\caption[]{a) A scatter plot showing the improvement in GAP score as a function of the models' diversity. b) shows the gap improvement (in \%) for different DNN pairs and c) shows the corresponding diversity score. In b),c) the type of DNN is identified as $\langle $feature$\rangle$ $\langle$ hid units$\rangle$ $\langle$ droupout$\rangle$, where for feature: M is mean, S is std. dev. and R means ROI.  }
 	\label{fig:gap_corr}
 \end{figure}

\subsection{Correlation-based Diversity Analysis}
The first analysis is based on Pearsons correlation coefficient defined as:
\[
R_{ij} = \frac{C_{ij}}{\sigma_i\sigma_j }
\]
where $i,j \in \{1,2,...,M\}$ and for classifiers indexed by $i$ and $j$, $R_{ij}$ is their correlation coefficient, $C_{ij}$ represents the covariance between these 2 classifiers and $\sigma_i,\sigma_j$ their respective output prediction standard deviations. Next, we find that one measure of diversity is: $1 - R_{ij}$, where if the correlation is minimal (i.e. 0), diversity is maximal and vice versa.

This method has the advantage of not requiring the classifier outputs to be binary, as is the case here.
When applied to the output predictions of the different classifiers in our ensembles, we find that a lower correlation is indicative of a greater improvement in the GAP20 score. This can be seen in Fig. \ref{fig:gap_corr}a). As shown there, we find that a a higher diversity score is highly correlated with an magnitude of improvement in the ensemble GAP score. Additional detail on the divergence scores and corresponding GAP20 improvement between pairs of DNNs can be seen in their respective heatmaps can be seen in Fig. \ref{fig:gap_corr}b,c.

Additionally, we can also use the diversity score to analyse the performance of overfitted models. This can be seen in Table \ref{table:overfit}. Here, we observe that when we allow a model to overfit past its highest validation score, this leads to an increase in diversity with other models. By ensembling overfitted models with lower individual scores, we actually observe that whilst this is detrimental to a single model's performance, it provides better improvement when incorporated into an ensemble.%an increase in the GAP20 improvement corresponds with a higher diversity score. This further suggests that when a model overfits, it improves diversity. Whilst this is detrimental to a single model's performance, it provides better improvement when incoporated into an ensemble.

\begin{table}[h!]
\centering
\begin{tabular}{|c| c | c|} 
 \hline
  & GAP Improvement & Diversity Score \\ 
 \hline \hline
 Peak Model & 0.579\% & 0.0322 \\ 
 \hline
 Overfitted Model& 0.671\% & 0.0340 \\ \hline
\end{tabular}
\bigskip
\caption{ Table showing the GAP improvement and diversity score for ensembles that use models with peak validation GAP20 or overfitted models with suboptimal GAP20 validation scores. }
\label{table:overfit}
\end{table}

\subsection{Generalised Diversity Measure-based Analysis}
Our second analysis is inspired by the Generalised Diversity Measure proposed by Partridge et al. \cite{Partridge}. In this measure, the authors propose that maximum diversity exists between two classifiers if, given an example, an error made by one classifier is accompanied by a correct classification of another classifier. 
% description of the difficult sets here
In order to obtain more insight into the improvements of classifier addition into an ensemble, we propose to analyse the performance of classifiers using ``wrong example sets''. 

Consider that each class has two sets of video examples, $N^+$ number of positive videos (label 1) and $N^-$ number of negative videos (label 0). Let these sets of videos be defined as $X^+ = \{x^+_1, x^+_2,...,x^+_{N^+}\}$ and $X^-= \{x^-_1, x^-_2,...,x^-_{N^-}\}$ respectively. Now, suppose we are given a classifier $h$, which can be an ensemble or single DNN. Correspondingly, the predictions given by $h$ on the different video sets are: $Y_h^+ = \{y^+_{h,1}, y^+_{h,2},...,y^+_{h,N^+}\}$ and $Y_h^- = \{y^-_{h,1}, y^-_{h,2},...,y^-_{h,N^-}\}$. 

We can now extract the set of videos that are considered ``wrong'' with respect to some threshold $\theta \in (0,1)$:
\begin{eqnarray*}
\varepsilon^+_{h,\theta} & = & \{i \in \{1,2,..., N^+\} : 1 - y^+_{h,i} \geq \theta\}\\
\varepsilon^-_{h,\theta} & = & \{i \in \{1,2,..., N^+\} : y^-_{h,i} \geq \theta\}
\end{eqnarray*}

The final set of ``wrong examples'' for classifier $h$ is:
\begin{equation}
\varepsilon_{h,\theta} = \varepsilon^+_{h,\theta} \cup \varepsilon^-_{h,\theta}
\label{eq:wrong_ex}
\end{equation}

We can now use the Eq. \ref{eq:wrong_ex} to analyse the effect of combining all of these classifiers together into an ensemble. In particular, we would like to discover if individual classifiers produce errors for {\em different} videos. If this were the case, when the classifiers are combined together, the erroneous predictions of individual classifiers can potentially be diluted by correct predictions from other classifiers.

To achieve this, first suppose we have an ensemble of $M$ classifiers: $H = h_1, h_2, ..., h_M$, and we assume that the errors for each classifiers are approximately equal. Next, the wrong-example sets are extracted using Eq. \ref{eq:wrong_ex} for each classifier, giving: $\varepsilon_{H,\theta} = \{\varepsilon_{h_1,\theta}, \varepsilon_{h_1,\theta}, ..., \varepsilon_{h_M,\theta}\}$. 

Now, consider the intersection of the sets in $\varepsilon_{H,\theta}$:
\[
\Upsilon_{H,\theta} = \bigcap^M_{i=1} \varepsilon_{h_i,\theta}
\]
The set of examples that fall into $\Upsilon_{H,\theta}$ are those that {\em all} the classifiers in the ensemble gave wrong predictions (w.r.t $\theta$) for. As such, this ensemble will not improve the predictions for any example in $\Upsilon_{H,\theta}$. Nonetheless, we find that the size of the set $\Upsilon_{H,\theta}$ either decreases or remains unchanged as we add new classifiers into the ensemble $H$.  

Additionally, we find that the union of sets in $\varepsilon_{H,\theta}$
\[
\Upsilon'_{H,\theta} = \bigcup^M_{i=1} \varepsilon_{h_i,\theta}
\]
represent the total unique videos that were wrongly classified by at least one classifier. However, examples in $\Upsilon'_{H,\theta}$ that are {\em not} in $\Upsilon_{H,\theta}$ will have an overall improved prediction in the ensemble.
\begin{figure}[t]
 	\centering
 	\includegraphics[width=\columnwidth]{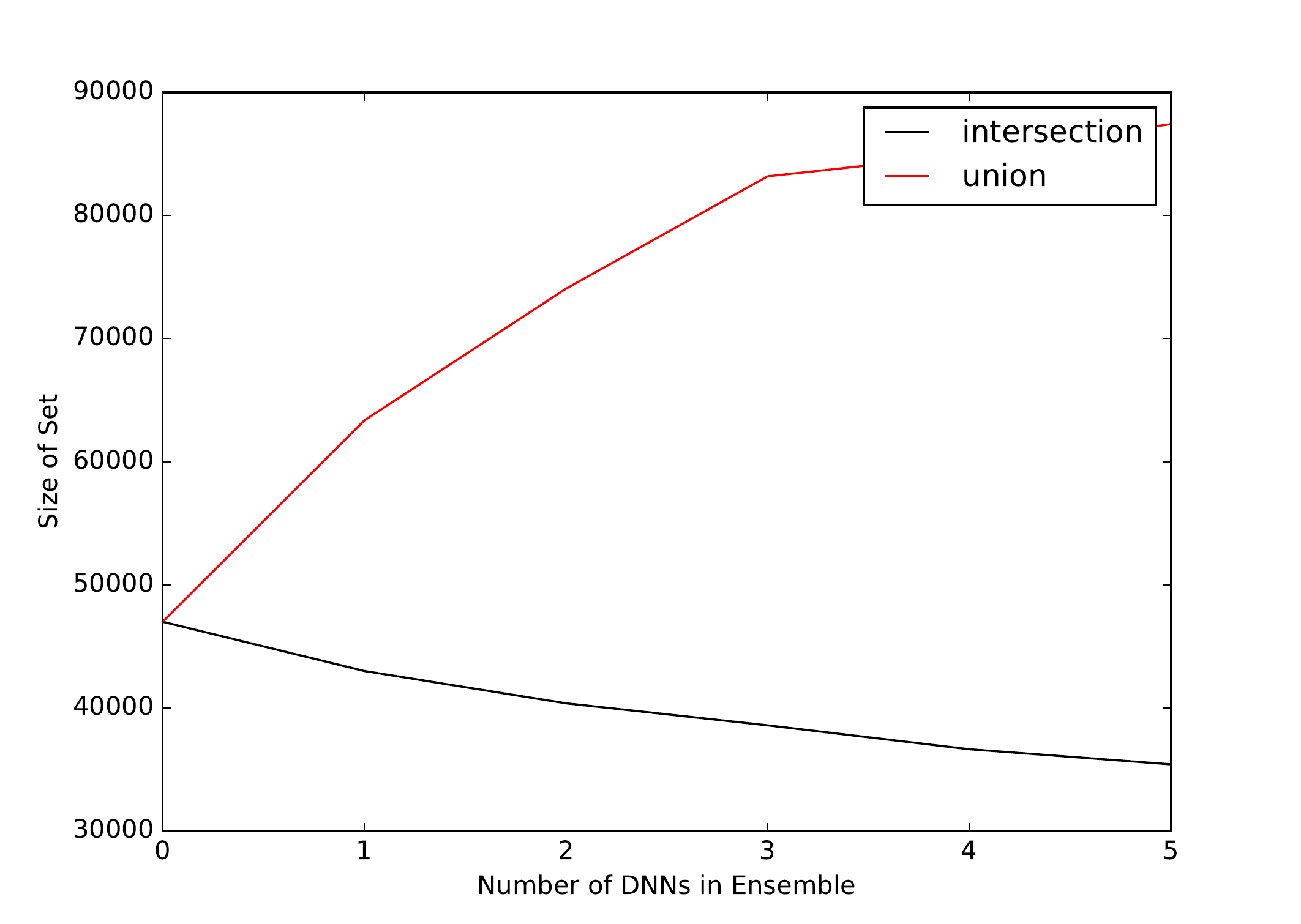}
 	\caption{ This graph shows the size of the sets representing the intersection of ``extremely wrong'' examples ($\theta = 0.9$) of individual classifiers in an ensemble, as more classifiers are added. Shown are also the size of the union of wrong examples that at least one classifier in the ensemble got significantly wrong. }
 	\label{fig:union_int}
\end{figure}

\begin{figure}[!h]
 	\centering
 	\includegraphics[width=\columnwidth]{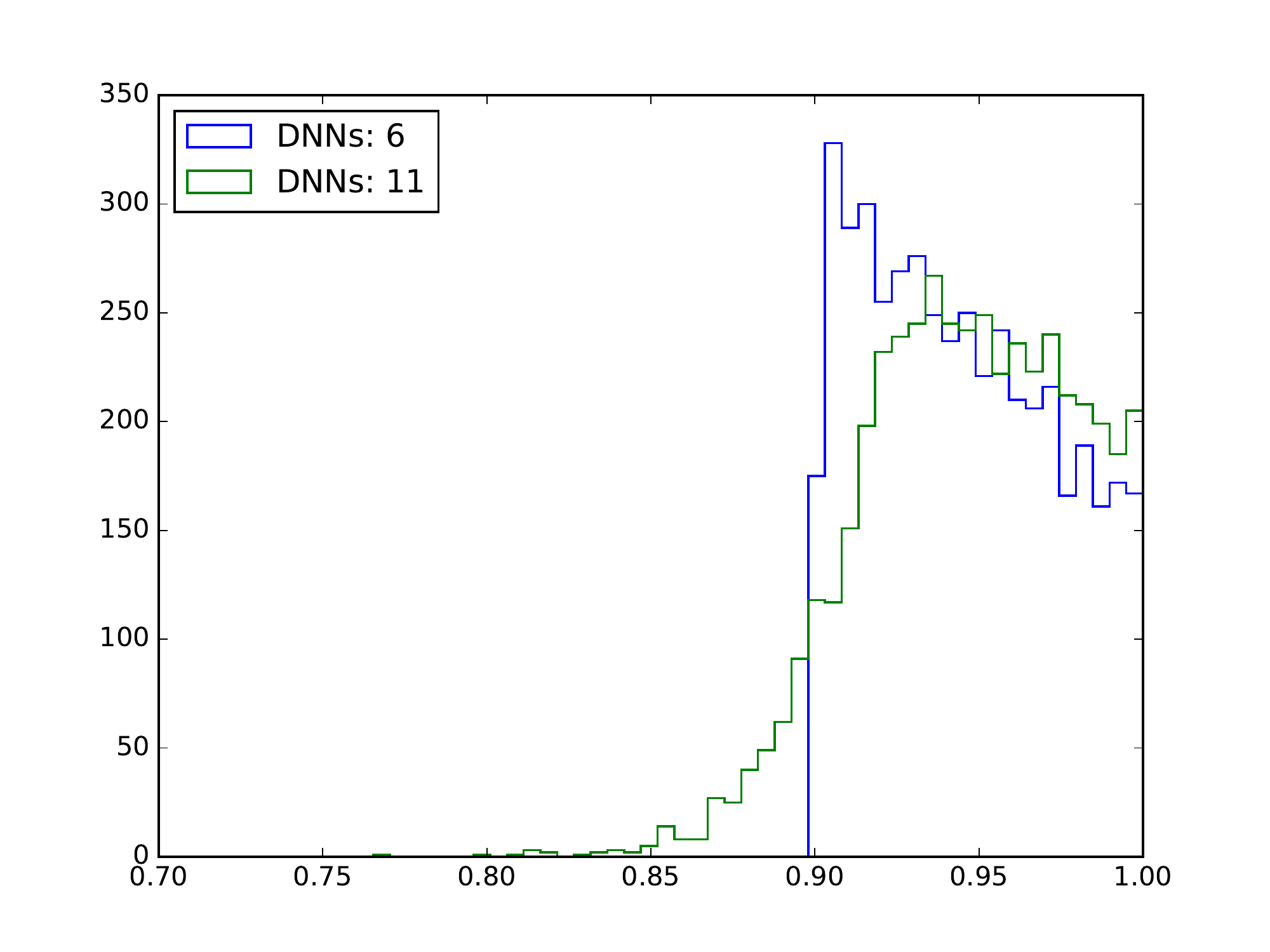}\\
 	(a)\\
 	\includegraphics[width=\columnwidth]{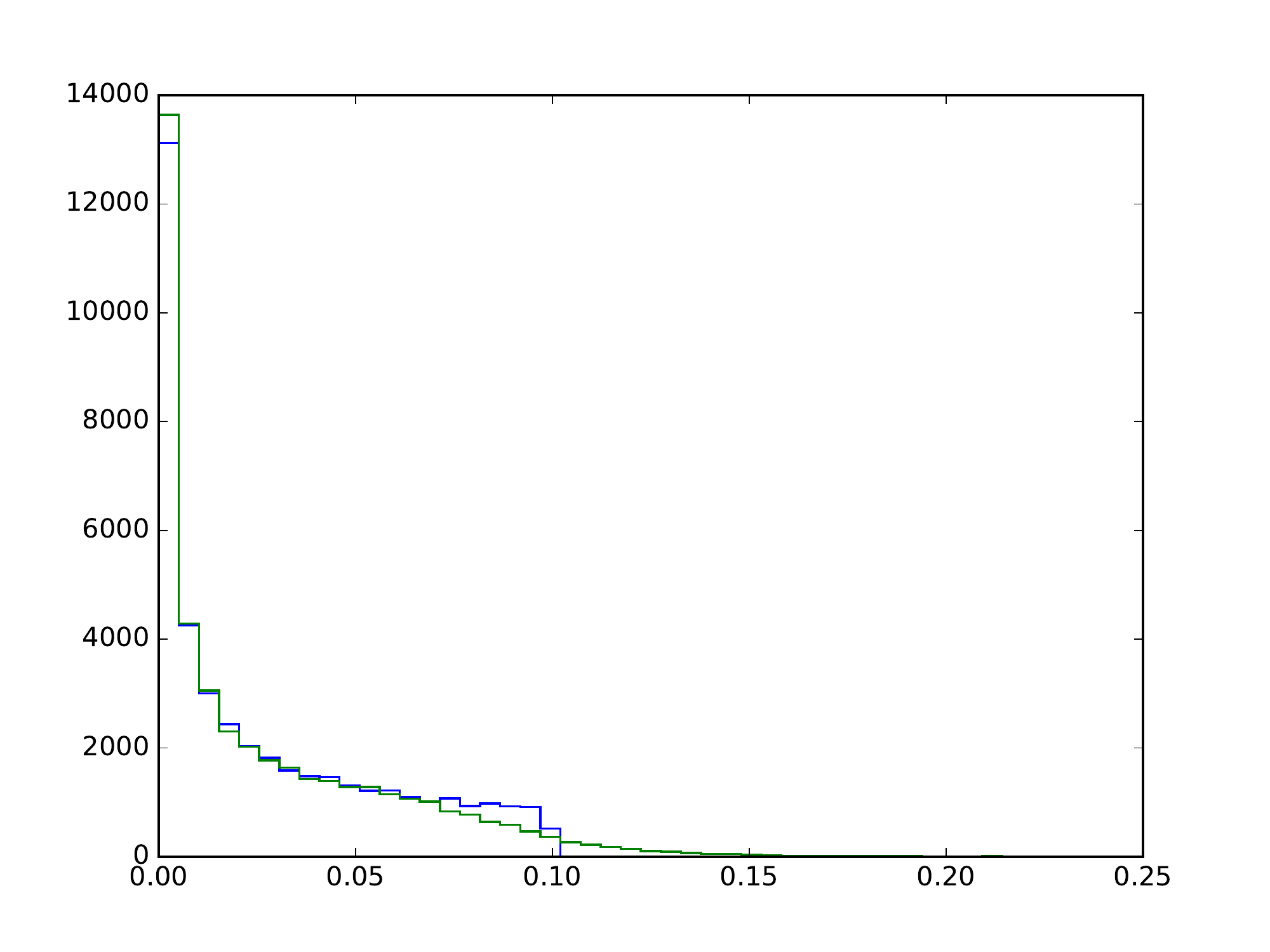}\\
 	(b)
 	\caption{ Shown here is how adding additional classifiers that are diverse into an ensemble diffuses the severity of wrong predictions. For clarity we have provided a zoomed-in view of prediction value histograms for wrong examples associated with very low predictions (+ve examples) (a) or very high predictions (-ve) examples (b). Shown are the histograms of examples with wrong predictions for two ensembles, one with 6 DNNs, and later when 5 more DNNs have been added.  }
 	\label{fig:diffuse}
\end{figure}
%\subsection{Final Ensemble Diversity}
Fig. \ref{fig:union_int} shows the size of $\Upsilon_{H,\theta}$ and $\Upsilon'_{H,\theta}$ as more classifiers are added to the ensembled used for this challenge, where $\theta = 0.9$. These represents extreme wrongly labelled videos. As such, these examples would have the greatest impact on decreasing the final GAP20 score. If these extreme mislabelling is due only to a small number of classifiers, then the ensemble should improve on their predictions (by means of accurate labelling from other classifiers).

Furthermore, if the above phenomena is occurring, we expect to see the intersection of ``wrong examples'' sets from individual classifiers decrease in size as we add more classifiers into the ensemble. This can indeed be seen in Fig. \ref{fig:union_int}. Here, we find that the number of examples that are wrongly labelled by all the individual classifiers in the ensemble steadily decreases as the ensemble size increases. This indicates that the individual classifiers of the ensemble each label different subsets of videos wrongly, suggesting that diversity is present. This in turn is results in a steady increase in the GAP score. 
An additional confirmation of the diversity is that the size of the union of wrong example sets is increasing. The classifiers in the ensemble all have approximately the same accuracy. That means their wrong example sets will be approximately the same size. Thus, their union will only expand in size if these examples are different.

Finally, we present results where we ``track'' the movement of extremely wrong predictions as we expand the ensemble size. We start by identifying the example videos that have prediction errors greater than $\theta = 0.9$. A histogram of their prediction scores is then built. We then obtain their predictions after a number of DNNs have been added in, and construct an updated histogram. The result of this is shown in Fig. \ref{fig:diffuse}. Here, the baseline-ensemble of 6 DNNs misclassified many examples and classes (approx. 42K), as can be observed in the blue histograms. However, after having added 5 additional DNNs that were found to be diverse, we find that many examples have smaller error scores, as shown in the green histograms. This will in turn result in the entries corresponding to these wrong predictions migrate further down the final sorted GAP list, thus improving the final GAP20 score.

\section{Conclusions}
\label{sec:conclusions}
% future work here

% conclusions here
In this paper we have investigated factors controlling DNN diversity in the context of the "Google Cloud and YouTube-8M Video Understanding Challenge". We have shown that diversity can be cultivated by using DNN different architectures. Surprisingly, we have also discovered diversity can be achieved through some unexpected means, such as model over-fitting and dropout variations. We have presented details of our overall solution to the video understanding problem, which ranked \#7 in the Kaggle competition (Yeti team - gold medal).

\section*{Acknowledgements}
The work in this paper was partially funded by Innovate UK under the iTravel project (Ref: 102811).

{\small
\bibliographystyle{ieee}
\bibliography{egbib}
}

\end{document}